# Contrastive Learning of Emoji-based Representations for Resource-Poor Languages

Nurendra Choudhary[*], Rajat Singh[*], Ishita Bindlish and Manish Shrivastava

Language Technologies Research Centre (LTRC)
Kohli Center on Intelligent Systems (KCIS)
International Institute of Information Technology, Hyderabad, India
{nurendra.choudhary,rajat.singh}@research.iiit.ac.in
ishita.bindlish@students.iiit.ac.in
m.shrivastava@iiit.ac.in

**Abstract** The introduction of emojis (or emoticons) in social media platforms has given the users an increased potential for expression. We propose a novel method called *Classification of Emojis using Siamese Network Architecture (CESNA)* to learn emoji-based representations of resource-poor languages by jointly training them with resource-rich languages using a siamese network.

CESNA model consists of twin Bi-directional Long Short-Term Memory Recurrent Neural Networks (Bi-LSTM RNN) with shared parameters joined by a contrastive loss function based on a similarity metric. The model learns the representations of resource-poor and resource-rich language in a common emoji space by using a similarity metric based on the emojis present in sentences from both languages. The model, hence, projects sentences with similar emojis closer to each other and the sentences with different emojis farther from one another. Experiments on large-scale Twitter datasets of resource-rich languages - English and Spanish and resource-poor languages - Hindi and Telugu reveal that CESNA outperforms the state-of-the-art emoji prediction approaches based on distributional semantics, semantic rules, lexicon lists and deep neural network representations without shared parameters.

## 1 Introduction

Social media continues to grow exponentially since its inception and has now become a forum filled with people's expression, opinions and sentiments. To better capture the text's sentimental context, users adopted *emojis*. Basically, emojis are special characters (or pictures) used to communicate context inexpressible by standard text. Emojis are ideograms and smileys used in electronic messages and web pages. Originally meaning pictograph, the word *emoji* comes from Japanese e (絵,picture) + moji (文字,character)[20]. Despite immense linguistic diversity, emojis and their definitions remain almost identical across all the major languages. Emojis capture a more mutually shared medium of communication, especially, in case of related cultures.

---

[*] These authors have contributed equally to this work.

| Lan | 😊 | ❤️ | 😍 | 😐 | 😠 | 😊 | 😘 | 😢 | 🤔 | 😶 | 🏠 | 😡 | 😄 | 😆 | 😉 | 💕 | 💙 | 💜 |
|---|---|---|---|---|---|---|---|---|---|---|---|---|---|---|---|---|---|---|
| Eng | 17.1 | 15.3 | 10.0 | 5.7 | 4.9 | 4.7 | 3.8 | 3.7 | 3.6 | 3.5 | 3.3 | 3.2 | 3.1 | 3.0 | 2.8 | 2.7 | 2.7 | 2.6 |
| Spa | 9.7 | 10.8 | 13.1 | 3.1 | 2.7 | 6.5 | 6.3 | 6.8 | 6.0 | 3.4 | 6.1 | 4.0 | 4.6 | 5.4 | 3.4 | 2.8 | 3.4 | 2.0 |
| Hin | 9.7 | 6.8 | 7.5 | 3.9 | 2.8 | 9.5 | 4.9 | 1.5 | 6.7 | 2.9 | 4.3 | 9.4 | 8.6 | 7.7 | 7.2 | 1.9 | 2.3 | 2.6 |
| Tel | 22.7 | 5.7 | 16.6 | 1.4 | 0.3 | 10.8 | 3.9 | 0.3 | 3.0 | 1.6 | 0.9 | 13.4 | 5.1 | 6.7 | 4.2 | 2.3 | 0.9 | 0.5 |

Table 1: Distribution of emojis in languages' tweets. [1]

A frequent usage of social media platforms is microblogging. These microblogs comprise of limited text with an emoji that represents the emotions related to that text. Hence, we establish a general correlation between the text and the emoji, where emoji is the corresponding text's tag in the microblog. Utilizing this aspect of emojis, we assert that sentences with similar corresponding emojis in different languages carry similar semantic features.

In this paper, we propose a novel unified framework called *Classification of Emojis using Siamese Network Architecture (CESNA)*. CESNA model consists of twin Bi-directional Long Short-Term Memory Recurrent Neural Networks (Bi-LSTM RNN) with shared parameters and a contrastive energy function, based on a similarity metric, joining them. The applied energy function suits discriminative training for energy-based models [14].

CESNA learns the shared model parameters and the similarity metric by minimizing the energy function connecting the twin networks. Parameter sharing and the similarity metric guarantee that, if the emoji of sentences on both the individual Bi-LSTM networks is same, then they are nearer to each other in the emoji space, else they are far away from each other. For example, the representations of *"The Big Bang Theory was funny 😄 "* and "चुनाव मजेदार थे 😄" (The elections were funny) should be nearer to each other than those of *"The Big Bang Theory was so funny today 😄 "* and "बिग बैंग थ्योरी उबाऊ था 😐" (Big Bang Theory was boring). The learned similarity metric is used to model the similarity between sentences of different languages into a common emoji space.

The rest of the paper is organized as follows. Section 2 presents the previous approaches to conquer the problem. Section 3 describes the evaluation dataset and section 4 describes the architecture of CESNA. Section 5 explains the training and testing phase of CESNA. Section 6 details the baselines. In section 7, the experimental set-up and results are presented. Finally, section 8 concludes the paper.

---

[1] The hearts in the table are of different colors. The most frequent one is *red*, the second most frequent one is *blue* and the last one is *purple heart*

## 2   Related Work

Distributional semantics [16] approach captures the overall sentence's semantic value but does not maintain information of the words' order. [17] assigns sentiment polarity to words or phrases. Polarity of its constituents assigns the score to the sentence. The information loss of the words' sequence leads to the wrong classification. e.g; In *"I am not happy"*, *"not"* carries a negative sentiment and *"happy"* has a positive sentiment. The combination gives a neutral sentiment, whereas the sentence is truly negative. Bag of n-grams limits the problem but does not eliminate it completely.

BiLSTM model[3] provides a solution to the problem of maintaining the sentence's sequence, by using recurrent neural network to embed sentences. They propose two types of embeddings based on words and characters. This approach presents an effective model for capturing the content and sequence in the sentence. However, the approach requires immense amount of data to train and hence will fail in case of languages with fewer resources.

Another line of research [12,2] utilizes rules and vocabulary of the languages to classify sentences. These techniques are highly accurate but susceptible to the problems of spelling errors and improper sentences. And these problems are very frequent in informal texts such as tweets. Also, in case of Hindi, [19] have trained a multinomial naive bayes model on annotated tweets to solve the problem.

Additionally, there have been efforts by researchers [18] to generate more annotated resources by utilizing available raw corpus. They employ the availability of different domains to construct a Multi-arm Active Transfer Learning (MATL) algorithm to label raw samples and continuously add them to the original dataset. Each step updates the algorithm's parameters using reinforcement learning with a reward function. The above approach works well for the domains considered in their work - sports, movies and politics. A formal grammar and vocabulary structure these domains, whereas, tweets do not follow this trend. Hence, the model is inapplicable to unstructured tweets. The new resources depend on the available resources' domain, which is risky, especially in the case of tweets that do not comply with any specific domain.

Most of the work done in the fields of emoji prediction and sentiment analysis is on major languages such as English or Spanish. Hence, the assumption in these approaches is the availability of immense data.

Usually, methods that require immutable words are ineffective. Applying languages' characters instead of words is a better approach. Given their proven effectiveness in [9,7,21,1,11,13,22], we use Bidirectional LSTMs (Bi-LSTMs) based on character n-grams. This approach produces embeddings based on the sequence of character n-grams, thus eliminating the problems of spelling mistakes and agglutination (in the case of some languages such as Telugu).

Although Bi-LSTMs manage mapping of sentences to an emoji space, we also require the distance between the sentences with the similar sentiment to be closer and the sentences with the different sentiment to be farther. For this reason, we use the architecture of siamese networks. This architecture possesses the capability of learning similarity from the given data without requiring specific information about the classes.

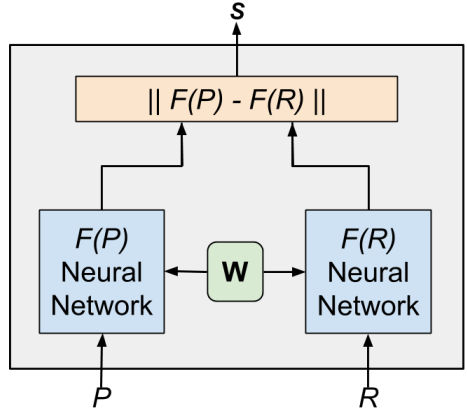

Figure 1: Siamese Network

### 2.1 Siamese Networks

[5] introduced siamese neural networks (shown in figure 1) to solve the problem of signature verification. Later, [6] applied the architecture with discriminative loss function for face verification. These networks also effectively enhance the quality of visual search [15,10]. Recently, [8] solved the problem of community question answering applying these networks.

Let $F(X)$ be the family of functions with parameters $W$. $F(X)$ is differentiable with respect to $W$. Siamese network seeks a value of the parameter $W$ such that the symmetric similarity metric is small if $X1$ and $X2$ belong to the same category, and large if they belong to different categories. The scalar energy function $S(R, P)$ that measures the emoji's relatedness between tweets of resource-poor ($P$) language and resource-rich ($R$) language can be defined as:

$$S(P, R) = ||F(P) - F(R)|| \tag{1}$$

In CESNA, the network takes tweets from both the languages as input. The loss function is minimized such that $S(P, R)$ is small if the $R$ and $P$ contain the same emoji and large otherwise.

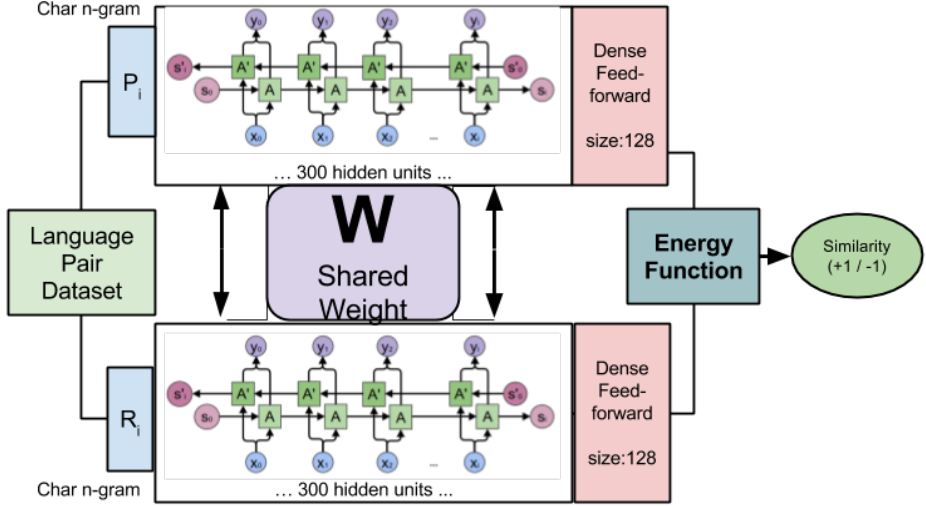

Figure 2: Architecture of CESNA

## 3 Dataset Creation

The twitter datasets for different languages are given below:

- **English:** Tweets from the ids given by [3]. The dataset consists of the tweets with 18 most frequent emojis, which is, 500,000 tweets.
- **Spanish:** Tweets containing the most frequent emojis present in English tweets, which is, 100,000 tweets.
- **Hindi:** Tweets containing the most frequent emojis present in English tweets, which is 15000.
- **Telugu:** Tweets containing the most frequent emojis present in English tweets, which is, 6000.

Table 1 demonstrates the distribution of the emojis in the above datasets.

## 4 Architecture of CESNA

As shown in figure 2, CESNA consists of a Bi-LSTM pair and a dense feed forward layer at the top. The Bi-LSTMs capture the sequence and constituents of the sentence and project them to a emoji space. We connect the yielded emoji vectors to a layer that measures similarity between them. The contrastive loss function combines the similarity measure and the label. Back-propogation through time computes the loss function's gradient with respect to the weights and biases shared by the sub-networks.

| Language | Hin | Tel | Eng | Spa |
|---|---|---|---|---|
| Char trigrams | 30849 | 21453 | 47924 | 42261 |
| Words | 41731 | 29298 | 72182 | 100171 |

Table 2: Number of Unique Character Trigrams and Words in the datasets

### 4.1 Primary Representation

Twitter Data consists of a lot of spelling errors, out-of-vocabulary words and word variations. The way of writing a word may also convey emotions (e.g; "Hi-iii" conveys a positive emotion whereas "Hi" is a neutral emotion). Hence, we use character trigrams to embed the sentence instead of using words. This approach takes care of the spelling errors and out-of-vocabulary words because a partial match exists in the character trigrams. Character trigrams take the information of all the word's inflections, thus, eliminating the problem of agglutination. This method, also, captures the information of different writing variations. Computational complexity is reduced as the number of words exceeds character trigrams. Table 2 shows the comparison between the number of unique words and unique trigrams in our case. The approach represents a sentence using a vector with number of dimensions equal to the number of unique character trigrams in the training dataset.

We input character-based term vectors of the resource-poor and resource-rich language's tweets and a label to the twin networks of CESNA. The label indicates whether the samples should be nearer or farther to each other in the emoji space. For positive samples (expected nearer in the emoji space), term vectors of tweets (one from resource-poor and one from resource-rich) with the same emoji are input to the twin networks. For negative samples (expected farther from each other in the emoji space), term vectors of tweets (one from resource-poor and one from resource-rich) with different emojis are input to the twin networks.

### 4.2 Bi-directional LSTM Network

We map each sentence-pair into $[p_i, r_i]$ such that $p_i \in \mathbb{R}^m$ and $r_i \in \mathbb{R}^n$, where $m$ and $n$ are the total number of character trigrams in the resource-poor language and the resource-rich language respectively.

Bi-LSTM model encodes the sentence twice, one in the original order (forward) of the sentence and one in the reverse order (backward). Back-propagation through time (BPTT) [4] calculates the weights for both the orders independently. The algorithm works in the same way as general back-propagation, except in this case the back-propagation occurs over all the hidden states of the unfolded timesteps.

We apply element-wise Rectified Linear Unit (ReLU) to the output encoding of the BiLSTM. ReLU is defined as: $f(x) = max(0, x)$. We choose ReLU here because it simplifies back-propagation, causes faster learning and avoids saturation.

The architecture's final dense feed forward layer converts the output of the ReLU layer into a fixed length vector $s \in \mathbb{R}^d$. In our architecture, we have empirically set the value of $d$ to 128. The overall model is formalized as:

$$s = max\{0, W[fw, bw] + b\} \tag{2}$$

where $W$ is a learned parameter matrix (weights), $fw$ is the forward LSTM encoding of the sentence, $bw$ is the backward LSTM encoding of the sentence, and $b$ is a bias term, then passed through an element-wise ReLU.

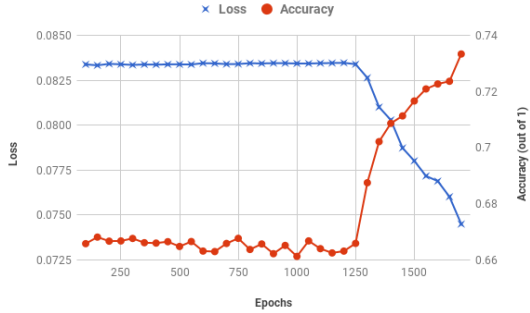

Figure 3: Loss and Accuracy vs Epochs

## 5 Training and Testing

We train CESNA on a tweet in resource-poor language with a tweet from resource-rich language to capture the similarity in the tweets' emojis. CESNA differs from the other deep learning counterparts due to its property of parameter sharing. Training the network with a shared set of parameters not only reduces the number of parameters (thus, save many computations) but also ensures that the sentences of both the languages are project into the same emoji space. We learn the shared network's parameters with the aim to minimize the distance between the tweets with the same emojis and maximize the distance between the tweets with different emojis.

Given an input $p_i, r_i$ where $p_i$ and $r_i$ are tweets from resource-poor and resource-rich languages respectively and a label $y_i \in \{-1, 1\}$, the loss function is defined as:

$$l(p_i, r_i) = \begin{cases} 1 - cos(p_i, r_i), & y = 1; \\ max(0, cos(p_i, r_i) - m), & y = -1; \end{cases} \tag{3}$$

where $m$ is the margin by which dissimilar pairs should move away from each other. It varies between 0 to 1. We minimize the loss function such that pair of tweets with the label 1 (same emoji) project nearer to each other and pair of tweets with the label -1 (different emoji) project farther from each other in the emoji space. The model trains by minimizing the overall loss function in a batch. The objective is to minimize:

$$L(\Lambda) = \sum_{(p_i, r_i) \in C \cup C'} l(p_i, r_i) \quad (4)$$

where $C$ contains the batch of same emoji tweet pairs and $C'$ contains the batch of different emoji tweet pairs. Back-propagation through time (BPTT) updates the parameters shared by the Bi-LSTM sub-networks.

| Dataset Pair | 5 | | | 10 | | | 18 | | |
|---|---|---|---|---|---|---|---|---|---|
| | P | R | F1 | P | R | F1 | P | R | F1 |
| Eng-Hin | 0.68 | 0.70 | 0.69 | 0.52 | 0.56 | 0.54 | 0.46 | 0.43 | 0.44 |
| Eng-Tel | 0.63 | 0.66 | 0.64 | 0.48 | 0.43 | 0.45 | 0.42 | 0.39 | 0.40 |
| Eng-Spa | 0.71 | 0.72 | 0.71 | 0.58 | 0.59 | 0.58 | 0.42 | 0.42 | 0.42 |
| Hin-Tel | 0.54 | 0.58 | 0.56 | 0.45 | 0.47 | 0.46 | 0.39 | 0.33 | 0.35 |
| Eng-Eng | 0.74 | 0.73 | 0.73 | 0.62 | 0.60 | 0.61 | 0.49 | 0.54 | 0.51 |

Table 3: Comparison between different Dataset Pairs for 5,10,18 emojis (classes). P,R,F1 are Precision, Recall and F-scores of the models respectively.

For testing, we randomly sample a certain number (100 in our case) of tweets for each emoji $R_{emoji}$ from the language corpus with higher amount of data. For every input, we then apply the trained model to get the similarity between the input and all corresponding $R_{emoji}$. The $R_{emoji}$ with the most number of matches with the input is finally selected as the correct emoji.

In case the testing data of both resource-rich and resource-poor languages do not contain emojis, we use the abundant resources of one language to construct a state-of-the-art emoji prediction model [3] and then utilize it to aid the resource-poor language's prediction.

## 6  Baselines

The approaches vary based on the language in consideration. Hence, we accordingly define the baselines below. English, Japanese and Spanish enjoy the highest share of data on Twitter[2]. We consider English and Spanish because of their script and typological similarity (both are Subject-Verb-Object). The baselines considered for resource-rich languages are:

- **Average Skip-gram Vectors (ASV):** We train a Word2Vec skip-gram model [16] on a corpus of 65 million raw (unannotated) tweets in English and 20 million raw tweets in Spanish. Word2Vec provides a vector for each word. We average the words' vectors in the tweet to get the vector for the sentence. So, each sentence vector is defined as:

$$V_s = \frac{\sum_{w \in W_s} V_w}{|W_s|} \quad (5)$$

---

[2] The Many Tongues of Twitter - MIT Technology Review

| Method | 5 | | | 10 | | | 18 | | |
|---|---|---|---|---|---|---|---|---|---|
| | P | R | F1 | P | R | F1 | P | R | F1 |
| ASV | 0.59 | 0.60 | 0.59 | 0.44 | 0.47 | 0.45 | 0.32 | 0.34 | 0.35 |
| Bi-LSTM (W) | 0.61 | 0.61 | 0.61 | 0.45 | 0.45 | 0.45 | 0.34 | 0.36 | 0.35 |
| Bi-LSTM (C) | 0.63 | 0.63 | 0.63 | 0.48 | 0.47 | 0.47 | 0.42 | 0.39 | 0.40 |
| DSC-T (RF) | 0.34 | 0.35 | 0.34 | 0.31 | 0.32 | 0.31 | 0.24 | 0.23 | 0.23 |
| MNB-H | 0.45 | 0.49 | 0.46 | 0.42 | 0.43 | 0.42 | 0.38 | 0.36 | 0.37 |
| CESNA (Eng-Eng) | **0.74** | **0.73** | **0.73** | **0.62** | **0.60** | **0.61** | **0.49** | **0.54** | **0.51** |
| CESNA (Hin-Eng) | **0.68** | **0.70** | **0.69** | **0.52** | **0.56** | **0.54** | **0.46** | **0.43** | **0.44** |
| CESNA (Tel-Eng) | **0.63** | **0.66** | **0.64** | **0.49** | **0.47** | **0.48** | **0.41** | **0.44** | **0.42** |

Table 4: Comparison with the Baselines. ASV is Average Skipgram Vectors, Bi-LSTM (W) and Bi-LSTM (C) refer to Word and Character based Bi-LSTM models. They are baselines for English and compare to CESNA (Eng-Eng). DSC-T is Domain Specific Classifier for Telugu and compares to CESNA (Tel-Eng). MNB-H refers to Multinomial Bayes Model for Hindi and compares to CESNA (Hin-Eng).P,R,F1 are the Precision, Recall and F-scores respectively.

Where $V_s$ is vector of the sentence $s$, $W_s$ is the set of words and $V_w$ is the vector of word $w$. After obtaining each message's embedding, we train an L2-regularized logistic regression, (with $\epsilon$ equal to 0.001).

– **Bidirectional LSTM (Bi-LSTM):** There are two approaches - word based and character based Bi-LSTM embeddings. We model the architecture as described in [3]. We use the same design as a part of our model, which is explained in Section 4.2.

Hindi and Telugu are the $3^{rd}$ and $17^{th}$ most spoken language in the world respectively. But they hold a relatively low share of twitter data. The major reason is that the speakers of Hindi and Telugu on Twitter primarily use the transliterated form of their respective language. The baselines for these languages are:

– **Domain Specific Classifier (Telugu) (DSC-T):** We train a Word2Vec model on a corpus of 700,000 raw Telugu sentences provided by Indian Languages Corpora Initiative (ILCI). We train a Random Forest (RF) and Support Vector Machines (SVM) classifier (given by [17]) on the Telugu Twitter dataset to structure our baseline for Telugu language.
– **Multinomial Naive Bayes (Hindi) (MNB-H):** We train a multinomial naive bayes model (given by [19]) on the Hindi Tweets dataset to form our baseline for Hindi language.

## 7 Experiments and Evaluation

In order to study the comparison between CESNA and the previous models, we performed an array of experiments. In the first experiment (Section 7.1), we analyze the model for varying language pairs and make a comparison between them. In the second experiment (Section 7.2), we compare our model against previous approaches in the problem. In the third experiment (Section 7.3), we train our architecture on clusters of emojis instead of unique ones.

### 7.1 Experiments for different language pairs

The experiment is a classification task. We take the English and Hindi Twitter datasets (Eng-Hin) and align each Hindi tweet with English tweets of the same emoji (positive samples) and label them 1. Similarly, we also randomly sample the same number of English tweets with different emoji (negative samples) for each Hindi tweet and label them -1.

We perform the experiment thrice taking 5 most frequent emojis (5 classes), 10 most frequent emojis (10 classes) and all the emojis (18 classes) in Hindi. Similarly, we repeat the experiment for English-Telugu (Eng-Tel) dataset pair, English-Spanish (Eng-Spa) dataset pair, English-English (Eng-Eng) dataset pair and Hindi-Telugu (Hin-Tel) dataset pair, taking 5 most frequent emojis (5 classes), 10 most frequent emojis (10 classes) and all the emojis in the language with lesser resource respectively for each case. The results of the experiments are given in Table 3.

### 7.2 Comparison with the Baselines

In this experiment, we compare our model against the baselines (defined in section 6). We defined the baselines for resource-rich languages on English. So, we perform contrastive learning of our model using data made by aligning each English tweet with a set of positive English tweet samples (with the same emoji) with label 1 and a set of negative English tweet samples (with different emoji) of the same size with label -1.

In the case of resource-poor languages, i.e. Hindi and Telugu, we perform contrastive learning of our model using data made by aligning each of the resource-poor language (Hindi and Telugu) tweet with a set of positive English tweet samples (with the same emoji) with label 1 and a set of negative English tweet samples (with different emoji) of the same size with label -1.

### 7.3 Clustering based approach with CESNA

In our previous experiment (Section 7.1), we observed that in several test scenarios the tweet is incorrectly classified to its nearest neighbor in the semantic space of emojis. We observe that the emojis form clusters in the semantic map. These clusters reduce multiple unique emojis to a single class for this experiment. So, we finally arrive at three clusters (table 5).

---

[3] The emojis in the *heart* cluster are of different colors. The first one is *red*, second one is *purple* and third one is *blue* in color.

Table 5: Clustering the emojis to a single emoji[3]

| Language Pair | P | R | F1 |
|---|---|---|---|
| CESNA (Eng-Eng) | 0.83 | 0.85 | 0.83 |
| CESNA (Hin-Eng) | 0.80 | 0.79 | 0.79 |
| CESNA (Tel-Eng) | 0.72 | 0.74 | 0.73 |

Table 6: Results after Clustering the Emojis

### 7.4 Evaluation of the Experiments

We observe from table 3 that the best overall results for multilingual emoji classification is the English-Spanish pair. This is due to the English-Spanish pair containing the maximum number of tweet pairs. We also note from figure 3 that with increasing number of epochs, the accuracy and overall performance considerably increases.

We also find that multiple times a tweet classifies into a related class. e.g; A tweet of class 💜 (*purple heart emoji*) is classified into a more frequent emoji ❤ (*red heart emoji*). To verify this behavior, we conducted another experiment in section 7.3 to approach this from the perspective of emojis' clustered classes. The results (given in table 6) demonstrate that fewer classes lead to better accuracy. This reduction in the number leads to a more even distribution of classes in the data. The drawback of this approach, though, is the loss of information about emojis. Hence, it only benefits when such data loss is acceptable.

From table 4, we observe that CESNA outperforms the state-of-the-art approaches significantly, especially in the case of resource-poor languages. Interestingly, table 4 also shows that using shared parameters (Siamese Networks) instead of a single Bi-LSTM network leads to an improvement in performance. CESNA learns representation, specifically, for the task of emoji-based classification. It also leverages the relatively resource-rich language for the improvement in the resource-poor language's accuracy.

## 8 Conclusions

In this paper, we proposed CESNA for emoji prediction of resource-poor languages which solves the problem by projecting the resource-poor language and resource-rich language in the same emoji space. CESNA employs twin Bidirectional LSTM networks with shared parameters to capture an emoji-based representation of the sentences. These emoji-based representations in conjunction with a similarity metric group sentences with similar emoji together.

A clustering based approach used in conjunction with CESNA boosts the performance of overall emoji prediction further. Experiments conducted on different

Twitter datasets revealed that CESNA outperforms the current state-of-the-art approaches significantly.

In future, we would apply the current model on more applications based on learning similarity like question-answering, conversation systems and semantic similarity. Though, of course, the presence and impact of emojis would be limited in other areas. Also, we believe that there is a good case for integration of attention-based models in the subnetworks.